\documentclass[10pt,letterpaper]{article}
\usepackage{spconf}
\usepackage[T1]{fontenc}
\usepackage{amsmath,amssymb,booktabs,array}
\usepackage{cite,graphicx,etoolbox}
\usepackage[hidelinks]{hyperref}
\usepackage{enumitem}
\makeatletter
\patchcmd{\@maketitle}{\large \bf \@title}
  {\fontsize{14}{16}\selectfont\bfseries \@title}{}
  {\PackageError{icassp-format}{Title font patch failed}{Use the supplied spconf.sty.}}
\makeatother

\newcommand{\dog}{\mathrm{dog}}

\title{WHEN DOES THE CONCEPT OF ``DOG'' EMERGE IN AN AUDIO LLM?}
\name{Zhe Wang, Shiqi Liu, Ruiyun Zhong, Tiechong Zhu,
Yihua Tan\sthanks{\fontsize{9}{10.8}\selectfont
Corresponding author: Yihua Tan.}}

\address{Huazhong University of Science and Technology,
Wuhan, China}
\begin{document}
\maketitle
\begin{abstract}
Multimodal large language models answer audio questions,
but how they represent auditory semantics and use them in
decisions remains unclear, limiting our understanding of
response formation. We study dog barking in Qwen2.5-Omni-7B using Jacobian lens
(J-lens) readout and directional interventions.
We define the dog direction as a J-lens-derived hidden-state
vector associated with dog; adding or removing its component
modulates dog-related information.
We find this information decodable without dog/bark prompt
cues or animal-identification requirements. Directional interventions change response tendencies and some final
answers, with effects concentrated in late-layer states
immediately before generation across species classification,
vocalization classification, and sound description.
The dog direction shows no comparable advantage over controls
in animal/other classification.
These results provide causal-intervention evidence that the
dog direction affects output scores in a task-dependent manner,
most consistently at L22 and L24 immediately before generation.
\end{abstract}

\section{Introduction}

Multimodal large language models can recognize sounds, answer audio-related questions, and generate open-ended descriptions~\cite{xu2025qwen}. Answering different questions about the same audio requires combining auditory information with task requirements. Yet final answers alone do not reveal how semantic information is represented internally or when it contributes to responses. Linking internal representations to response behavior can help explain how auditory information supports language generation and task-specific decisions. Existing methods examine both the decodability and the functional role of internal representations. Linear probes test whether information is decodable, but decodability alone does not establish its use during answer generation~\cite{alain2016probes,hewitt2019probes}. The Jacobian lens (J-lens) uses cross-layer Jacobian relationships to map intermediate states into a common late-layer representation space, enabling readout of verbalizable representations through the model's own vocabulary~\cite{gurnee2026verbalizable}. Concept ablation and activation interventions test the influence of representations on outputs by modifying internal states~\cite{elazar2021amnesic,turner2023activation}. Readout identifies decodable information; interventions test its influence on outputs. Research on verbalizable representations also examines how
internal information supports different
computations~\cite{gurnee2026verbalizable}.
This motivates testing whether the same auditory concept
influences responses across tasks.

\begin{figure}[!t]
    \centering
    \includegraphics[width=\columnwidth]{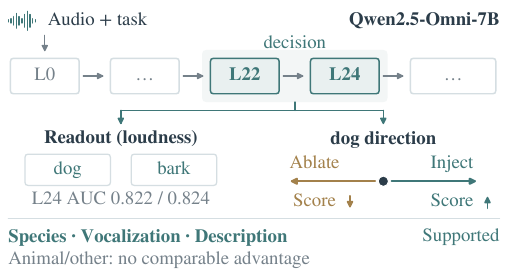}
    \caption{Overview of the main findings: concept readout,
    late-layer interventions, and task-dependent effects.
    AUCs refer to the loudness task.}
    \label{fig:core_findings}
\end{figure}

However, how nonlinguistic sounds are encoded in verbalizable
representations and how these representations contribute to
responses across tasks remain insufficiently understood. Not naming a sound source
during loudness judgment does not imply that source
information is absent, while candidate labels may influence
category readout. We therefore test readout without
target-category cues or identification requirements.
Output-score changes may reflect nonspecific perturbations
and need not alter final answers, motivating control
interventions and full-response generation. Comparisons
across layers, token positions, and tasks assess where these
effects occur and whether they extend across tasks.

We study dog barking in Qwen2.5-Omni-7B through dog (source),
bark (vocalization type), and animal (superordinate category),
with cat vocalizations providing a related comparison.
We first validate a text-fitted J-lens on audio inputs.
We define the dog direction as a J-lens-derived vector in
hidden-state space associated with the token dog;
interventions modify states along this vector.
We then follow a readout--intervention--localization sequence
to examine dog-related information and its influence on
responses, using norm-matched controls in full-response
and localization experiments. Our contributions are:

\begin{enumerate}
\item We combine J-lens readout and directional interventions
to examine dog-related representations. We find this
information decodable without dog/bark prompt cues or
animal-identification requirements.

\item Interventions along the dog direction change species-classification scores and some final answers. For non-dog inputs, dog-direction injection yields higher answer-transition rates than norm-matched controls at the compared dose, providing causal evidence that this direction influences answer generation.

\item The dog direction most consistently influences responses in layers 22–-24 at the last input position before answer generation. In species classification, vocalization classification, and sound description, removing/adding the dog component respectively lowers/raises dog-related answer scores. This advantage does not extend to superordinate animal/other judgments, indicating task-selective effects. 
Figure~\ref{fig:core_findings} provides an overview of our
main findings.
\end{enumerate}

\section{Methods}

We follow a readout--intervention--localization sequence: we first examine dog-related information in tasks that do not require animal identification, then measure response changes under intervention, and finally compare these effects across layers, positions, and tasks.

\subsection{Model and Data}

We use the Thinker component of Qwen2.5-Omni-7B, whose language decoder contains 28 Transformer blocks, numbered 0--27. We denote layer $\ell$ by L$\ell$ and read and intervene on residual-stream states at selected block outputs. Dog-barking recordings are drawn from DogSpeak~\cite{lekhak2025dogspeak}, cat vocalizations come from a dataset we assembled, and other sounds are primarily drawn from ESC-50~\cite{piczak2015dataset} and UrbanSound8K~\cite{salamon2014dataset}.

The concept-ranking experiment uses 50 dog-barking  recordings and 50 non-dog recordings to analyze full-vocab\-u\-lary ranks and their changes across layers. Readout under tasks unrelated to animal identification and subsequent interventions use the same set of 600 recordings, comprising 200 dog recordings, 200 cat recordings, and 200 recordings of other sounds. Cat vocalizations provide a closely related animal-vocalization control. The other-sound category contains 190 environmental sounds without animal vocalizations and 10 vocalizations from animals other than dogs and cats.

\subsection{J-lens Mapping and Concept Readout}

J-lens maps states from L0, L4, L8, L12, L16, L20, L22, and L24 to the penultimate layer L26, to limit possible final-block readout artifacts~\cite{gurnee2026verbalizable}. The mappings are estimated from 1,000 text prompts and validated on held-out prompts and audio inputs without audio-domain refitting. For $h_{\ell,t}\in\mathbb R^d$, the fixed mean cross-layer
Jacobian $W_\ell$ and bias vector $b_\ell$ give
\begin{equation}
\tilde h_{\ell,t}=W_\ell h_{\ell,t}+b_\ell.
\label{eq:jlens}
\end{equation}
For concept token $k$ with output-head vector $v_k$, the scaled cosine readout score is
\begin{equation}
s_k(h)=\sqrt d\,
\frac{\tilde h_{\ell,t}^{\top}v_k}
{\|\tilde h_{\ell,t}\|_2\|v_k\|_2}.
\label{eq:score}
\end{equation}
The factor $\sqrt d$ leaves sample ordering and AUC unchanged. We assess decodability by comparing dog and non-dog recordings. We use single-token directions for concepts such as dog and bark; intervention directions are derived from the output head and J-lens mapping.

Full-vocabulary ranks use J-lens output-head logits and are summarized by the within-group median across recordings at each layer and position; AUCs use Eq.~\eqref{eq:score}. Uninterpretable tokens are excluded only during semantic interpretation. We also compare $s_{\rm bark}-s_{\rm noise}$ and $(s_\dog+s_{\rm bark})-(s_{\rm sound}+s_{\rm noise})$ using equal weights without fitting an additional classifier.

\subsection{Intervention Methods and Controls}
We test dog ablation, dog injection, and dog--cat swapping
against control interventions.

Let $h$ and $h'$ denote the pre- and post-intervention states.
We construct the unnormalized concept direction as
$d_{\ell,k}=W_\ell^\top v_k$, where $W_\ell$ is the frozen
cross-layer Jacobian matrix used in the J-lens mapping
and $v_k$ is the output-head vector for token $k$.
Ablation and injection use the unit direction
$u_{\ell,k}=d_{\ell,k}/\|d_{\ell,k}\|_2$,
whereas swapping uses the unnormalized directions.

For $k=\mathrm{dog}$, ablation on dog recordings and injection
on non-dog recordings are defined as

\begin{equation}
\begin{aligned}
h'&=h-\beta\langle h,u_{\ell,\mathrm{dog}}\rangle
u_{\ell,\mathrm{dog}}
&&\text{(ablation)},\\
h'&=h+\alpha\|h\|_2u_{\ell,\mathrm{dog}}
&&\text{(injection)}.
\end{aligned}
\label{eq:interventions}
\end{equation}
Here, $\beta=1$ removes the projection onto the dog direction,
whereas $\beta=2$ reverses it; $\alpha$ specifies the injected
norm relative to $\|h\|_2$. For swapping, we compute the
minimum-norm least-squares coefficients of $h$ using the
unnormalized directions $d_{\ell,\mathrm{dog}}$ and
$d_{\ell,\mathrm{cat}}$, exchange the coefficients, and add
the resulting reconstructed-state difference to $h$ with
strength $\gamma$.

Ablation and injection use vehicle and random orthogonal
directions as controls. Layerwise injections match target
and control perturbation norms, whereas layerwise ablations
compare trends without strict norm matching. Full-response
generation and localization strictly match perturbation norms;
full-response generation additionally uses sound--vehicle
swapping as a swap control.

\subsection{Tasks and Experimental Procedure}

We use seven tasks:
\begin{enumerate}[
    label=(\arabic*),
    leftmargin=1.8em,
    labelsep=0.4em,
    topsep=2pt,
    itemsep=0pt,
    parsep=0pt,
    partopsep=0pt
]
\item \emph{Fixed response:}
output \texttt{ready} regardless of audio content.

\item \emph{Loudness:}
judge whether the sound is quiet or loud.

\item \emph{Event count:}
judge whether one or multiple sound events occur.

\item \emph{Species:}
classify the sound source as dog, cat, or other.

\item \emph{Vocalization type:}
identify the vocalization type rather than the species.

\item \emph{Sound description:}
briefly describe the audio in natural language.

\item \emph{Animal/other:}
classify the source as animal or non-animal.
\end{enumerate}

\textbf{Task readout.}
We compare tasks (1)--(4) at L24 decision, the
pre-generation newline token. Tasks (1)--(3) require
no animal identification and omit dog/bark from prompts.

\textbf{Layerwise interventions and generation.}
For species classification, we intervene at decision in
L0, L8, L16, L20, L22, and L24 with $\beta=1$,
$\alpha=0.03$, and $\gamma=1$.
Full-response experiments at L24 vary
$\beta\in\{1,2\}$, $\alpha\in\{0.03,0.06,0.10\}$,
and $\gamma\in\{0.5,1,1.5\}$.
Both experiments inject into other-sound recordings.

\textbf{Localization.}
For tasks (4)--(7), we compare L16, L20, L22, and L24
at audio middle (midpoint audio-content token),
audio end (last audio-content token),
post-audio (first task-text token after
\texttt{audio\_eos}), and decision.
Ablation uses dog recordings; injection uses cat and
other-sound recordings for tasks (4)--(6) and
environmental sounds for task (7).
Each intervention modifies one block-output state
at one token and one layer during prefill.
\subsection{Evaluation Metrics and Statistical Tests}

We assess decodability using AUC~\cite{hanley1982meaning}
and $\Delta_{\mathrm{read}}=\bar s_D-\bar s_N$, where
$\bar s_D$ and $\bar s_N$ are the mean readout scores
from Eq.~\eqref{eq:score} for dog-barking and non-dog
recordings, respectively.

Let $z_k$ denote the model's output-head logit for token $k$
at the first answer-generation step, before
softmax~\cite{goodfellow2016deep}. Layerwise ablation/injection
and sound description use $z_{\mathrm{dog}}$; swapping uses
$z_{\mathrm{dog}}-z_{\mathrm{cat}}$. Species, vocalization,
and animal/other localization use log-odds
$\log[p_k/\sum_{j\ne k}p_j]$ for dog, bark, and animal,
respectively~\cite{hosmer2013applied}, where $p_k$ is the
first-step token probability and the sum covers the remaining
task candidates. Let $\Delta_I$ be the mean paired score
change, intervention minus baseline. Answer-transition rates
are the fractions switching to a specified answer among
recordings with the designated baseline answer~\cite{davis2014matched}.

For localization, $\Delta_{\mathrm{dog}}$ is the mean
score change from baseline under dog-direction intervention;
$\Delta_{\mathrm{ctrl}}$ averages vehicle and random orthogonal
control changes within each recording, then across recordings.
We define
\begin{equation}
E=
\begin{cases}
\Delta_{\mathrm{ctrl}}-\Delta_{\mathrm{dog}},
& \text{ablation},\\
\Delta_{\mathrm{dog}}-\Delta_{\mathrm{ctrl}},
& \text{injection}.
\end{cases}
\label{eq:effect_difference}
\end{equation}
Thus, $E>0$ means a larger decrease under ablation or
a larger increase under injection relative to controls.
Magnitudes are not compared across different metric scales.

We estimate 95\% confidence intervals (CIs) using 10,000
bootstrap resamples, sampling recordings with replacement
while preserving each recording's baseline--intervention
pairings~\cite{efron1993bootstrap}. Two-sided sign-flip permutation tests yield $p$-values;
Benjamini--Hochberg (BH) correction across 240 localization
tests gives adjusted values denoted by $q$~\cite{benjamini1995controlling}.
Support requires the target score to decrease under ablation
or increase under injection. Its advantage over controls
must also satisfy $E>0$, a 95\% CI entirely above zero,
and $q<0.05$.

\section{Experiments and Results}
\label{sec:results}
\subsection{Mapping Validation and Concept Readout}

\textbf{Mapping validation.}
Under audio inputs, cosine similarity between mapped and actual L26 states increases with layer depth, reaching 0.820 at L24 with 98.5\% output-head top-1 agreement. This supports late-layer readout, while substantial mapping error limits early-layer interpretation.

\textbf{Concept readout.}
Across eight source layers and different input stages, we evaluate full-vocab\-u\-lary ranks and readout AUCs for 50 dog-barking and 50 non-dog recordings. Table~\ref{tab:concept_ranks} shows that dog distinguishes the groups (AUC 0.937) at L24, while bark remains informative (AUC 0.872) despite ranking far below the top vocabulary items. In this experiment,
$(s_{\mathrm{dog}}+s_{\mathrm{bark}})
-(s_{\mathrm{sound}}+s_{\mathrm{noise}})$
achieves a higher AUC than $s_{\mathrm{bark}}-s_{\mathrm{noise}}$.

\begin{table}[t]
\centering
\caption{Median full-vocab\-u\-lary concept ranks at the L24 decision position (smaller ranks indicate higher vocabulary positions). AUC is computed from normalized readout scores.}
\label{tab:concept_ranks}
\small
\begin{tabular}{@{}lrrr@{}}
\toprule
Concept & Dog rank & Non-dog rank & AUC \\
\midrule
animal  & 1.0    & 58.5    & 0.786 \\
dog     & 55.0   & 53,028.5 & 0.937 \\
vehicle & 695.0  & 31.0     & 0.166 \\
bark    & 964.5  & 19,653.5 & 0.872 \\
\bottomrule
\end{tabular}
\end{table}

\subsection{Concept Readout in Tasks Unrelated to Animal Identification}

Using 200 dog and 400 non-dog recordings, we test whether dog and bark scores at L24 decision distinguish the groups without an animal-iden\-ti\-fi\-ca\-tion requirement. Fixed-re\-sponse, loudness, and event-count prompts contain neither dog nor bark; species classification provides a comparison. Table~\ref{tab:task_readout} reports selected tasks: both concepts remain discriminative under fixed-response and loudness prompts, supporting readout without explicit category cues.

\begin{table}[t]
\centering
\caption{Separability of dog-barking and non-dog recordings in selected tasks. $\Delta_{\rm read}$ is the mean readout-score difference; the fixed response is ready.}
\label{tab:task_readout}
\small
\setlength{\tabcolsep}{3pt}
\begin{tabular}{@{}llrrr@{}}
\toprule
Task & Token & $\Delta_{\rm read}$ & 95\% CI & AUC \\
\midrule
Fixed response & dog & 1.270 & $[1.133,1.403]$ & 0.873 \\
Fixed response & bark & 0.554 & $[0.494,0.613]$ & 0.865 \\
Loudness & dog & 0.544 & $[0.475,0.613]$ & 0.822 \\
Loudness & bark & 0.249 & $[0.216,0.281]$ & 0.824 \\
Species & dog & 3.659 & $[3.508,3.800]$ & 0.986 \\
Species & bark & 0.881 & $[0.827,0.934]$ & 0.957 \\
\bottomrule
\end{tabular}
\end{table}

\subsection{Effects of Directional Interventions on Output Scores and Final Generated Answers}

\textbf{Layerwise interventions.}
To test whether the dog direction influences species judgments, we apply three operations at the decision position in each of six selected layers: ablating the dog component for 200 dog-barking recordings, injecting the dog direction for 200 recordings of other sounds, and exchanging the coordinates along the dog and cat directions for dog-barking and cat-vocalization samples. Each intervention modifies only one token at one layer while keeping the audio and prompt unchanged. We compare first-step output scores before and after intervention for the same sample.

Intervention effects are close to zero at L0 and L8, increase from L20 onward, and are most pronounced at L24. At L24, ablation decreases the dog logit by an average of 2.806, whereas injection increases it by an average of 1.279. Swapping changes the dog--cat logit difference by $-4.799$ for dog samples and $+4.396$ for cat samples. These changes follow the expected intervention directions, indicating that intervening along concept directions can influence output scores, with more pronounced effects in later layers.

\textbf{Final answers.}
At L24 decision, we compare label orders A (dog, cat, other) and B (cat, dog, other). Under A, increasing $\beta$ from 1 to 2 raises dog$\rightarrow$non-dog transitions from 14/191 to 53/191. Injection at $\alpha=0.06$ yields 20/180 other$\rightarrow$dog transitions, versus 0/180 and 1/180 for norm-matched vehicle and random orthogonal controls. Swapping at $\gamma=1.5$ yields 93/191 dog$\rightarrow$cat transitions but only 1/190 cat$\rightarrow$dog transitions. Denominators count recordings with the respective baseline answers. At L24 decision, $\beta=2$ yields more
dog$\rightarrow$non-dog transitions than $\beta=1$;
dog injection at $\alpha=0.06$ yields more
other$\rightarrow$dog transitions than norm-matched controls.

\subsection{Intervention Localization and Task Differences}

At L22 and L24 decision, six of eight task--operation comparisons meet the support criteria, versus at most two at other tested positions. These six comparisons cover ablation and injection in species classification, vocalization classification, and sound description; animal/other classification shows no corresponding advantage (Table~\ref{tab:localization_effects}).
Under single-token interventions, these results support task-dependent effects at late-layer decision positions. Limited baseline performance in vocalization classification and sound description restricts cross-task conclusions primarily to continuous scores; final-answer evidence mainly comes from species classification.

\begin{table}[t]
\centering
\caption{Mean effect differences $E$ at L24 decision with norm-matched controls. $\dagger$: meets the prespecified support criteria. Metric scales differ across tasks.}
\label{tab:localization_effects}
\small
\setlength{\tabcolsep}{5pt}
\begin{tabular}{@{}llr@{}}
\toprule
Task & Operation & $E$ (L24) \\
\midrule
Species & Ablation & $0.767^{\dagger}$ \\
Species & Injection & $1.185^{\dagger}$ \\
Vocalization type & Ablation & $0.119^{\dagger}$ \\
Vocalization type & Injection & $0.214^{\dagger}$ \\
Sound description & Ablation & $2.502^{\dagger}$ \\
Sound description & Injection & $2.432^{\dagger}$ \\
Animal/other & Ablation & $-0.117$ \\
Animal/other & Injection & $-0.180$ \\
\bottomrule
\end{tabular}
\end{table}

\section{Conclusion}
We study dog barking in Qwen2.5-Omni-7B through semantic
readout and directional interventions. First, dog-related information remains decodable without
dog/bark prompt cues or animal-identification requirements.
Furthermore, dog-direction interventions change output scores
and some final species-classification answers.
Finally, across tested layers and positions, these effects
are most consistent at L22 and L24 immediately before
answer generation.
The direction produces larger score changes in the expected
direction than matched controls in species classification,
vocalization classification, and sound description, and shows
no corresponding advantage in animal/other judgments.
These results support a task-dependent role of the dog
direction in shaping model responses.
Combining readout with controlled interventions links
hidden-state information to answer changes.
Comparisons across tasks, layers, and token positions clarify
where these effects are most consistent and how they depend
on task requirements.
\clearpage

\bibliographystyle{IEEEbib}
\bibliography{Mybib}
\end{document}